\documentclass[letterpaper, 10 pt, conference]{ieeeconf}  

\IEEEoverridecommandlockouts                              

\usepackage{array,booktabs,verbatim,amsmath}
\usepackage{graphicx}  
\usepackage{array}     
\usepackage{booktabs}  
\usepackage{tabularx}
\usepackage{times}
\usepackage{latexsym}
\usepackage{booktabs}
\usepackage{array} 
\makeatother
\usepackage{tcolorbox}  
\usepackage{mdframed}

\newtcolorbox{mybox}{colback=red!5!white,colframe=red!75!black}
\usepackage{framed}

\tcbuselibrary{skins}  
\usepackage{float}
\usepackage[T1]{fontenc}

\usepackage[utf8]{inputenc}

\usepackage{microtype}
\usepackage{listings} 
\usepackage{inconsolata}
\usepackage{multirow}

\usepackage{graphicx}

\title{\LARGE \bf
Towards Robust Retrieval-Augmented Generation Based on Knowledge Graph: A Comparative Analysis
}

\author{Hazem Amamou$^{1, 3}$, Stéphane Gagnon$^{2}$, Alan Davoust$^{2, 3}$ and Anderson R. Avila$^{1, 3}$\\\\
$^1$Institut national de la recherche scientifique (INRS-EMT), Montréal, Québec, Canada\\%
$^2$Université du Québec en Outaouais, Gatineau, Québec, Canada\\%
$^3$INRS-UQO Mixed Research Unit on Cybersecurity, Gatineau, Québec, Canada%
}

\begin{document}

\maketitle
\thispagestyle{empty}
\pagestyle{empty}

\begin{abstract}
Retrieval-Augmented Generation (RAG) was first introduced to enhance the capabilities of Large Language Models (LLMs) beyond their encoded-prior knowledge. This is achieved by providing LLMs with an external source of knowledge, which helps to reduce factual hallucinations and enables the access to new information, typically not available during their pretraining phase. Despite its benefits, there is an increasing concern with the impact of inconsistent retrieved information towards LLMs' responses. Hence, the Retrieval-Augmented Generation Benchmark (RGB) was introduced as a new testbed for RAG evaluation, meant to assess the robustness of LLMs towards inconsistency in the retrieved information. In this work, we use the RGB corpus to evaluate LLMs in four scenarios: (1) noise robustness; (2) information integration; (3) negative rejection; and (4) counterfactual robustness. We perform a comparative analysis between the RAG baseline defined by the RGB and variations of GraphRAG, which is a RAG system based on a Knowledge Graph (KG) and developed to retrieve relevant information from large documents. We tested GraphRAG under three customization to improve its robustness. Our approach demonstrates improvements compared to the RGB baseline, providing insights on how to design more reliable RAG systems, tailored for real-world scenarios.

\end{abstract}

\section{INTRODUCTION}

Large Language Models (LLMs) account for millions of direct users. Platforms such as OpenAI, Google Cloud AI and Anthropic provide chatbots that can support individuals seeking for general advices, help with content generation, coding or text analysis. If we consider indirect use of AI-powered tools, such as text editors, coding environment (e.g., IDE) and customer support systems, the number of users scales up to the order of billions. All this has become possible due to recent advancements in LLMs. Nevertheless, these models still face serious problems such as factual hallucination \cite{ji2023}, outdated knowledge \cite{wang2023knowledge}, and absence of domain-specific expertise \cite{shen2023}. Output errors from LLMs can negatively impact a large number of people. To overcome these limitations and mitigate the consequences to the end-user, Retrieval-Augmented Generation (RAG) has emerged \cite{lewis2020retrieval}. It provides external knowledge via information retrieval performed on knowledge corpora \cite{thakur2024knowing}, which is meant to enhance the accuracy and reliability of LLM responses \cite{gao2024retrieval}. 

With the vast amount of real-time information available on the Internet, it is common for chatbots (e.g., ChatGPT) to retrieve documents by searching the web. In such scenarios, the retrieved documents might contain inaccurate information leading to unreliable responses. Therefore, there have been an increasing interest not just in understanding the impact of retrieval-augmented generation on LLMs, but also in mitigating potential performance degradation due to inconsistent content retrieval. Here, we highlight three potential limitations of the RAG framework: (1) retrieved texts may contain redundant or noisy information, increasing LLM confusion \cite{lewis2020retrieval}; (2) unstructured formats obscure explicit relationships between entities (e.g., temporal/causal links), hindering multi-hop reasoning \cite{lin2021trig}; and (3) contradiction detection requires costly cross-document analysis \cite{thorne2018fever}.

\begin{figure*}
    \centering
    \includegraphics[width=0.70\linewidth]{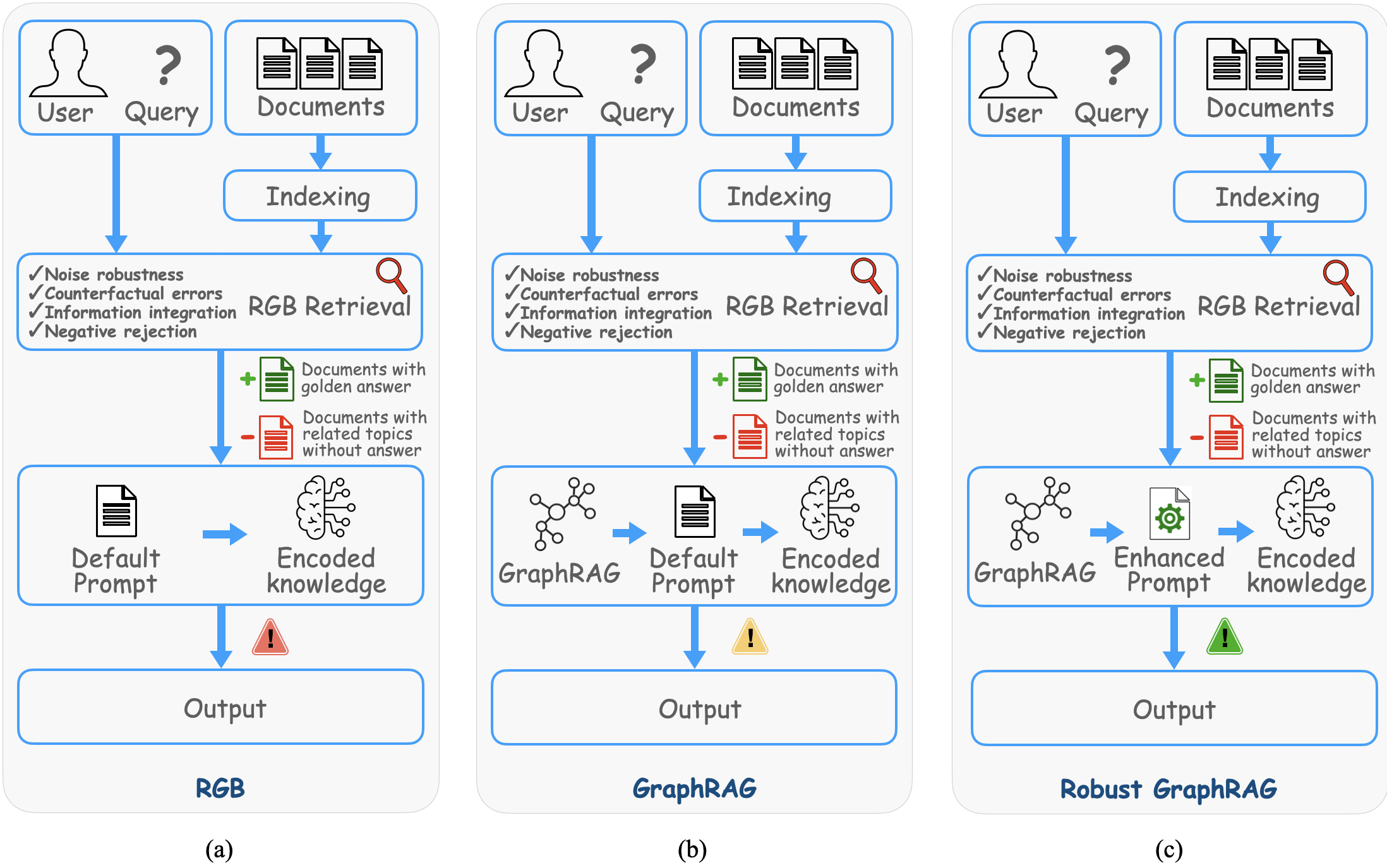}
    \caption{Comparative Framework of RGB Benchmark, GraphRAG, and RobustGraphRAG for Evaluating Robustness in Retrieval-Augmented Generation}
    \label{fig:robustgraphrag}
\end{figure*}


In such context, few benchmarks have been proposed to systematically evaluate LLM performances within the RAG framework. In this study, we adopt the Retrieval-Augmented Generation Benchmark (RGB) \cite{chen2024benchmarking} to evaluate four abilities required for assuring LLM robustness while relying on retrieved information from external knowledge. Thus, RAG systems are assessed based on \emph{noise robustness}, which is the ability to extract correct information from noisy documents; \emph{information integration}, which refers to the ability of LLMs to answer complex questions that require integrating answers from multiple documents; \emph{negative rejection}, which is the capability of rejecting to provide an answer when the retrieved documents contain irrelevant information and the LLM encoded-prior knowledge is not enough to respond the question; and \emph{counterfactual robustness}, which refers to the ability to identify and disregard incorrect or contradictory information in retrieved documents, thereby minimizing the propagation of factual errors in generated responses.

In this study, we address these limitations by exploring the combination of RAG with Knowledge Graph (KG). While RAG retrieves unstructured documents, which may contain irrelevant or redundant information, KG offers structured knowledge representation that can assist to improve the context provided to LLMs. This is specially relevant for factual queries directly targeting entities, relationships, claims or specific attributes. Our experiments are based on a specific framework, namely GraphRAG \cite{edge2024}, which is a graph-based approach that relies on community detection for handling more general questions while properly scaling for large texts. As in our experiments we found that directly applying GraphRAG to the RGB provides limited improvements compared to the baseline provided by the benchmark, we propose few customizations to GraphRAG, tailoring it to address the four issues presented in the RGB Benchmark. Figure \ref{fig:robustgraphrag} depicts the main scenarios explored in this work. Note that the information retrieval comprises positive (relevant and correct) and negative (irrelevant or noisy/incorrect) documents. Moreover, our solution requires not just some level of customization of GraphRAG, but also the use of an enhanced prompt. In general, results show that our approaches mitigate the impact of inconsistency in the retrieved context.

\section{Methodology}
\label{sec:methodology}

We start this section by presenting GraphRAG, which is the foundation for this study. We then provide details about the proposed customizations, which aimed to address the four challenges presented in the RAG benchmark, refer to as RGB.

\subsection{Preliminaries}

GraphRAG is a knowledge-graph-based approach for question answering \cite{edge2024local}. The framework was developed with the aim of being robust towards handling generic user questions and processing long documents with a large amount of texts. It builds an entity-based knowledge graph using LLM capabilities. The process is divided in two stages. The first stage is the derivation of entity knowledge graph, and the second is the generation of community summaries for groups of closely related entities. GraphRAG combines local and global summaries to attain a hierarchical description of an entire corpus. In summary, the pipeline of the framework consists of splitting all documents into text chunks, then prompting an LLM to extract entities and its relationships with their respective descriptions. The next step is to create summaries of concepts extracted by the LLM to then build a knowledge graph. This knowledge graph is used to create community summaries, with lower-level communities being used to generate summaries for higher-level ones. These community summaries are utilized to answer users' questions from different hierarchical levels.


\subsection{GraphRAG for LMM Robustness}

We introduce few customizations to improve the robustness of GraphRAG. We are particularly concern with scenarios where retrieved information convey noise, factual errors or simply cannot provide the answer. Note that GraphRAG was design to handle dependencies in large text corpus, while managing users generic queries. While it fits well with the challenge of information integration, it was not tailored with specific capabilities to mitigate the other issues addressed in this work. Thus, our assumption is that by introducing structured external knowledge into the retrieval-generation pipeline, we can enhance reasoning while mitigating issues related to inaccurate retrieved information. To achieve our objective, the following modifications are proposed:

\subsubsection{Document Indexing} In this step, documents are split into chunks, and entities, relationships and claims are identified to build the entity-based knowledge graph. Note that the RGB assumes that the retrieval was already performed. To reflect this scenario, we consider building the structured knowledge directly from the retrieved documents. Thus, the structured knowledge is based on both positive and negative documents provided by the RGB retrieval step. Four different approaches are proposed to improve the robustness of LLMs' responses. The first is referred to as GR$_{RGB}$ and represents the default prompt used in the RGB, but adapted to KG context, as shown in Figure \ref{fig:robustgraphrag}-b. The exact prompt used is provided in Figure \ref{fig:prompt_RGB} (see appendix). We also considered a variation of this approach, namely GR$_{def}$, where the prompt used for question answering in the original GraphRAG framework is adopted. To address the challenges presented by the RGB, we customized these prompts for each task (see the appendix for detailed information). Additionally, we also investigated the influence of the use of the LLM- encoded -prior knowledge versus the exclusive use of retrieved information from the external knowledge. This was meant to evaluate how much the internal knowledge can contribute to hallucinations, i.e., the attempt by the LLMs to rely on its internal knowledge even when it lacks the correct information to answer user questions. Our enhanced prompt is depicted in Figure \ref{fig:robustgraphrag}-c, and referred to as GR$_{ext}$, for the exclusive use of external knowledge, and GR$_{comb}$, for the combination of encoded and external knowledge.
\begin{figure}

\begin{mdframed}[backgroundcolor=blue!5, linecolor=blue!80!black]
$\textbf{GR$_{\textbf{RGB}}$ Contextual Prompt}$

~

You are an accurate and reliable AI assistant that can answer questions with the help of external documents. Please note that external documents may contain noisy or factually incorrect information. If the information in the document contains the correct answer, you will give an accurate answer. If the information in the document does not contain the answer, you will generate 'I cannot answer the question because of the insufficient information in documents.'. If there are inconsistencies with the facts in some of the documents, please generate the response 'There are factual errors in the provided documents.' and provide the correct answer.

~

STRUCTURED CONTEXT: \{KG context\} 

~
    
OUTPUT: \{answer\} 

~

\end{mdframed}
    \caption{GR$_{RGB}$ prompt presented to the LLM, based on the RGB system combined with structured knowledge.}
    \label{fig:prompt_RGB}
\end{figure}

\begin{figure}

\begin{mdframed}[backgroundcolor=blue!5, linecolor=blue!80!black]
$\textbf{GR$_{\textbf{ext}}$ Contextual Prompt}$

~

 You are a helpful assistant tasked with answering questions based on the provided external documents. The documents may contain noisy or irrelevant information. Your goal is to extract the necessary information to answer the question accurately, ignoring any irrelevant or noisy content.
    Instructions:
    1. **Focus on Relevant Information**: Carefully analyze the documents and extract only the information that directly answers the question.
    2. **Ignore Noise**: If a document contains irrelevant or noisy information, ignore it and focus on the parts that are relevant to the question.
    3. **Provide a Clear Answer**: If the answer is found in the documents, provide it clearly and concisely.
    4. **Reject if Necessary**: If the documents do not contain sufficient information to answer the question, respond with:
    *"I cannot answer the question due to insufficient information in the documents.

~

STRUCTURED CONTEXT: \{KG context\} 

~
    
OUTPUT: \{answer\} 

~

\end{mdframed}
    \caption{GR$_{ext}$ External-Only GraphRAG Prompt for Robust, Noise-Aware Answering}
    \label{fig:prompt_ext_NR}
\end{figure}

 \begin{table}
    \centering
    \small
    \caption{Descriptions of the structured knowledge based approaches assessed in our experiments.}
    \scalebox{0.99}{
    \begin{tabular}{lp{0.77\columnwidth}}
    \toprule
         Model & Description\\\midrule
         RGB & Standard RGB baseline\\
         GR$_{RGB}$& GraphRAG with RGB prompt\\
         GR$_{def}$& GraphRAG with its default prompt\\ 
         GR$_{ext}$& Customized GraphRAG with only external knowledge \\
         GR$_{comb}$& Customized GraphRAG with encoded and external knowledge\\
         \bottomrule
    \end{tabular}}
    \label{tab:model}
\end{table}


\begin{figure*}
    \centering
    \includegraphics[width=0.8\linewidth]{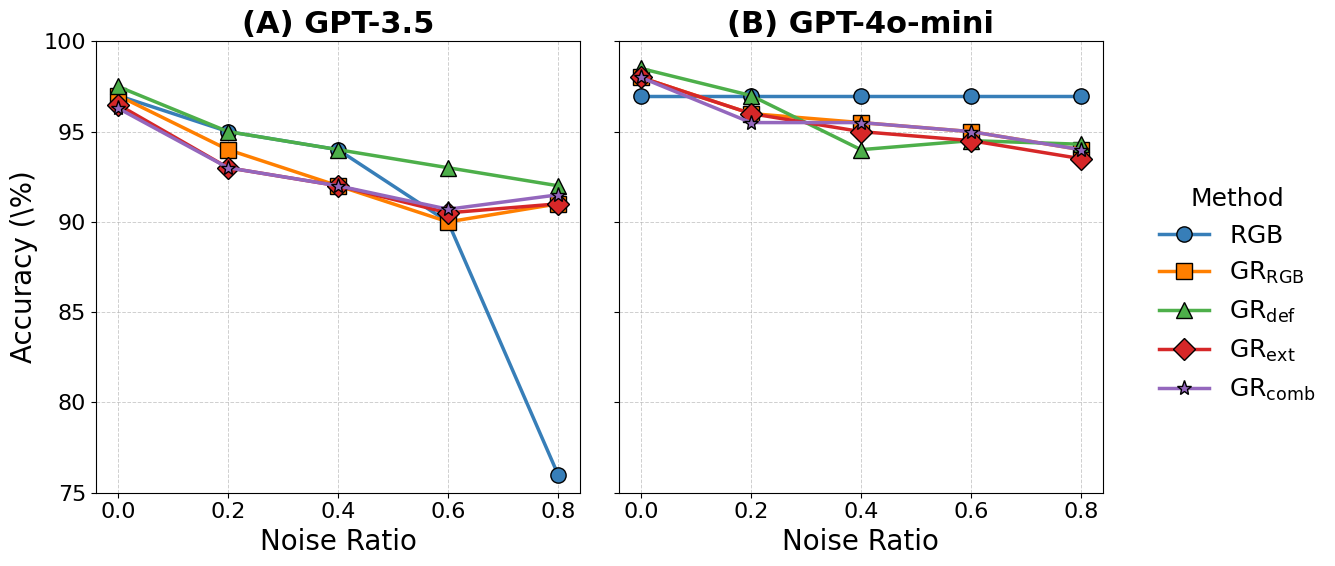}
    \caption{Comparative analysis for the  noise robustness task measured by accuracy (\%) for GPT-3.5 and GPT-4o-mini under varying noise ratios.}
    \label{fig:noise}
\end{figure*}


\section{Experimental Setup}
\label{sec:setup}

\subsection{RGB Benchmark}
We utilize the Retrieval-Augmented Generation Benchmark (RGB) \cite{chen2024benchmarking} to evaluate LLMs' robustness towards inconsistancy in the retrieved documents. We assess whether LLMs can effectively use external documents to generate reasonable answers, comparing the RGB baseline with four structured knowledge configurations. The RGB benchmark is specifically designed to assess the capabilities of RAG systems across four specific tasks:  \emph{negation rejection}, which occurs when a model attempts to provide an answer despite lacking sufficient knowledge to respond accurately;  \emph{information integration}, which refers to when large language models (LLMs) face challenges in combining information from multiple documents;  \emph{counterfactual robustness}, which relates to scenarios where the model encounters contradictory information; and  \emph{noise robustness}, which addresses situations where the retrieved information includes irrelevant or inaccurate content. We selected the RGB baseline as it is the only benchmark specifically designed to evaluate RAG robustness across the four targeted tasks (noise robustness, information integration, etc.). Unlike general QA benchmarks (e.g., HotpotQA \cite{yang2018hotpotqa}), RGB provides controlled noise injection and counterfactual document sets, enabling direct measurement of LLM sensitivity to inconsistent retrievals. 

\subsection{Evaluation Metrics}
We adopt the same evaluation metrics used in the RGB benchmark \cite{chen2024benchmarking}. The metrics are specifically tailored to each task in the benchmark. \emph{Accuracy} (ACC) is used to assess performance in the information integration and noise robustness experiments. \emph{Rejection Rate} is considered to test negative rejection capability. Thus, if provided with just noisy documents, LLMs are expected to refuse to generate a response. \emph{Error Detection Rate} (ED) is used to evaluate model's ability to identify factual errors in documents, and is applied specifically for the counterfactual robustness task. \emph{Error Correction Rate} (CR) evaluates the ability of the model to give the correct answer after detecting errors in the counterfactual robustness task. 


\subsection{Experimental Settings}

In order to index the documents provided in the RGB retrieval step, we utilize GPT-4o-mini. For generating answers, on the other hand, we tested both GPT-4o-mini and GPT-3.5. All experiments use identical hyperparameters to ensure fair model comparisons. For the knowledge graph construction, a confidence score of 0.7 for entities was adopted which refers to the minimum threshold required for an entity to be included in the knowledge graph. The Leiden algorithm for community detection was used with a resolution parameter of 1.0. We kept the number of retrieved documents fixed to 5 for all experiments. In our experiments, GraphRAG processes each query independently by constructing a knowledge graph from only the retrieved documents relevant to that query. Since the KG is built dynamically for each query, the overall corpus size does not affect latency or accuracy. This design ensures consistent performance regardless of external corpus scale, as indexing and community detection are applied only to the minimal document set required for the task. Future work may explore batch processing of larger document sets for efficiency in multi-query scenarios.

\section{Experimental Results}
\label{sec:results}
We evaluate the customized GraphRAG on the four tasks of the RGB benchmark—noise robustness, counterfactual robustness, information integration, and negative rejection. Results are presented for both GPT-4o-mini and GPT-3.5, highlighting the effect of model complexity on performance across different RAG approaches.

\subsection{Noise Robustness}

Figure~\ref{fig:noise} shows the accuracy for the noise robustness task across different noise ratios, representing retrieved documents with different levels of inaccurate or irrelevant information. We evaluate the performance of GPT-3.5 and GPT-4o-mini, with the latter showing more robustness towards all conditions. For example, while the baseline provided by the RGB benchmark showed no decay in terms of accuracy for GPT-4o-mini, GPT-3.5 was severely impacted by noise, achieving the lowest accuracy when the noise level was set to 0.8. This suggests that the encoded knowledge of more complex models, such as GPT-4o-mini, plays an important role in handling noisy information present in the retrieved context. In general, GPT-3.5 was more affected by noise, as the performance of all configurations decayed as the noise ratio increased, with the most robust approach being the GR$_{def}$. This configuration represents the default prompt used by the GraphRAG framework. This configuration is tailored to process long documents from large texts, which explains its performance in Figure 3 for GPT-3.5. The second best configuration is the GR$_{comb}$, followed closely by the GR$_{ext}$. Thus, the combination of encoded and external knowledge seems to be the best approach for less complex models. For the GPT-4o-mini, GraphRAG achieved better results only for the lowest noise ratio of 0.2. This highlights that integrating knowledge graph retrieval, especially with enhanced prompt design, can be an alternative for more complex models when the external knowledge is reliable and contains less inaccurate information. For less reliable sources, combining encoded and external knowledge significantly improves the performance of less complex models. These gains are particularly evident with GPT-3.5, where GraphRAG yields a clear advantage. Overall, incorporating knowledge graphs in retrieval-augmented generation settings can greatly enhance LLM robustness to noisy inputs, especially for smaller or less capable models.


\subsection{Counterfactual Robustness Evaluation}

\begin{table}
\caption{Comparative analysis for the counterfactual robustness task. ACC represents the accuracy (\%) of LLMs without external knowledge. ACC$_{doc}$ is the accuracy (\%) of  LLMs with counterfactual documents. ED* is the error detection rate and CR is the error correction rate.}
\centering
\small
\setlength{\tabcolsep}{6.0pt} 
\renewcommand*{\arraystretch}{0.70}

\setlength{\tabcolsep}{6pt} 

\centering
\begin{tabular}{llcccc}
\toprule\vspace{0.05cm}
& & ACC & Acc\textsubscript{doc} & ED* & CR \\
\midrule
\multirow{6}{*}{\rotatebox{90}{\footnotesize GPT3.5}} 
 & RGB & 93.00 & 9.00 & 7.00 & 57.14  \\\vspace{0.05cm}
 & GR$_{RGB}$ & N/A & 1.00 & 4.00 & 25.00  \\\vspace{0.05cm}
 & GR$_{def}$ & N/A & 7.00 & 60.60 & 95.00  \\\vspace{0.05cm}
 & GR$_{ext}$ & N/A & 3.00 & 15.15 & 13.33  \\\vspace{0.2cm}
 & GR$_{comb}$ & N/A & \textbf{24.24} & \textbf{94.94} & \textbf{95.74}  \\

\multirow{6}{*}{\rotatebox{90}{\footnotesize GPT4omini}} \vspace{0.05cm}
 & RGB & 96.00 & \textbf{71.00} & 77.00 & \textbf{100.00}  \\\vspace{0.05cm}
 & GR$_{RGB}$ & N/A & 11.11 & 16.16 & 81.25 \\\vspace{0.05cm}
 & GR$_{def}$ & N/A & 8.00 & 62.26 & 98.38  \\\vspace{0.05cm}
 & GR$_{ext}$ & N/A & 0.00 & 62.62 & 0.00  \\
 & GR$_{comb}$ & N/A & 35.35 & \textbf{100.00} & 85.85 \\
\bottomrule
\end{tabular}
\label{tab:counterfactual}
\label{tab:counterfactual_main}

\end{table}

Table \ref{tab:counterfactual_main} provides a comparative performance for the counterfactual robustness task. We can observe that GPT-3.5 is more impacted by documents containing factual errors, see the ACC$_{doc}$ column, being outperformed by GPT-4o-mini in all configurations, except for the GR$_{ext}$, i.e., when the external knowledge is exclusively used. When relying only on the external knowledge, GPT-4o-mini provides 0 \% accuracy, while GPT-3.5 achieves 3 \%. Note that there is a significant difference between the RGB baseline performance, which is 9 \% for GPT-3.5, and 71 \% for GPT-4o-mini. This corroborates with our findings in the previous section where the use of encoded knowledge by the more complex model led to better performance. The GraphRAG configurations were more relevant for GPT-3.5, with the GR$_{comb}$ being the least vulnerable to factual errors. It achieves 24.24 \% accuracy. For the GPT-4o-mini, the same configuration had a negative impact on the results compared to the accuracy achieved with the RGB baseline, 71 \%, although it outperformed all other GraphRAG configuration, achieving 35.35 \%. On the other hand, in terms of error detection (ED), the combination of the LLM encoded knowledge and external knowledge, GR$_{comb}$, led to the best results, achieving 94.94 \% and 100 \%, respectively, for GPT-3.5 and GPT-4o-mini, while the RGB benchmark achieved 7 \% and 77 \%. It shows that less complex models will face more difficulties to detect errors in the retrieved context, unless they utilize our customized GraphRAG. Regarding the correction rate (CR), the solution based on GPT-3.5, GR$_{comb}$, was successful in correcting 95.74 \% of the detected errors, whereas GPT-4o-mini achieved the best correction rate with the RGB benchmark. The configuration based on the default GraphRAG, GR$_{def}$, was efficient in detecting and correcting errors for both LLMs. The drawback of this configuration is that it provided very low ACC$_{doc}$. 

\begin{table}
\caption{Comparative analysis for the information integration task measured in terms of accuracy (\%) for different noise ratios..}
\centering
\small
\setlength{\tabcolsep}{6.0pt} 
\renewcommand*{\arraystretch}{0.70}

\setlength{\tabcolsep}{6pt} 

\centering
\begin{tabular}{llccc}
\toprule\vspace{0.05cm}
 &  & \multicolumn{3}{c}{Noise Ratio} \\
\cline{3-5}\noalign{\vskip 3pt}
& & 0.0 & 0.2 & 0.4 \\
\midrule

\multirow{6}{*}{\rotatebox{90}{\footnotesize GPT3.5}} 
 & RGB & 55.00 & 51.00 & 34.00 \\\vspace{0.05cm}
 & GR$_{RGB}$ & 83.83 & 79.00 & 71.00 \\\vspace{0.05cm}
 & GR$_{def}$ & 81.81 & 77.00 & 72.00 \\\vspace{0.05cm}
 & GR$_{ext}$ & \textbf{86.86} & \textbf{83.00} & \textbf{78.00} \\\vspace{0.2cm}
 & GR$_{comb}$ & \textbf{86.86} & \textbf{83.00} & 75.00 \\

\multirow{6}{*}{\rotatebox{90}{\footnotesize GPT4omini}} 
 & RGB & 82.00 & 79.00 & 80.00 \\c
 & GR$_{RGB}$ & 79.79 & 78.00 & 73.00 \\\vspace{0.05cm}
 & GR$_{def}$ & 86.86 & 85.00 & 82.00 \\\vspace{0.05cm}
 & GR$_{ext}$ & \textbf{87.87} & \textbf{86.00} & 81.00 \\\vspace{0.05cm}
 & GR$_{comb}$ & 86.86 & 84.00 & \textbf{83.00} \\

\bottomrule
\end{tabular}
\label{tab:integration}

\label{tab:integration_main}

\end{table}

\subsection{Information Integration Evaluation}
Table \ref{tab:integration_main} presents the results for the information integration task, which evaluates the model's ability to synthesize information from multiple documents to answer the user question. The performance is presented across three noise conditions: 0 \%, 20 \% and 40 \%. The RGB baseline exhibits significant sensitivity to noise for this task. GPT-3.5, for instance, has its accuracy decreased from 55 \%, at 0 \% noise, to 34 \%, at the highest noise level of 0.4. For GPT-4mini, on the other hand, the drop is less severe, going from 82.00 \% to 80 \%. These results suggest that GPT-4o-mini handles noise better than GPT-3.5 even with the baseline configuration. For the GraphRAG models, such as the GR$_{RGB}$, which uses RGB prompts, performance was always better compared to the baseline, for both LLMs. At 0 \% noise, the best performances are attained using the GR$_{ext}$ approach, that is, with the LLM relying only on the external knowledge. This was expected given that GraphRAG was developed to process information from long documents, which implies the ability to integrate relevant information that might be spread in different parts of a document to answer the user question. Note that all KG configurations are affected by noise, with a steady decline in accuracy as the noise ratio increases. This was again more severe for GPT-3.5. GR$_{def}$, which relies on default prompts, maintains higher stability, particularly in GPT-4o-mini, where the accuracy remains above 80 \% even at the highest noise level. Leveraging external knowledge (GR$_{ext}$), emerges as the most robust model for both LLMs, displaying minimal performance degradation across noise levels. GPT-3.5 achieves 86.86 \% accuracy when noise is not present, and decreases to 78.00 \%, at the noise ratio of 0.4, while GPT-4o-mini achieves slightly better results, dropping from 87.87 \%, with the absence of noise, to 81.00 \%, at the noise ratio of 0.4. GR$_{comb}$ which combines encoded and external knowledge, performs similarly to GR$_{def}$, although it exhibits slightly higher sensitivity to noise, particularly for GPT-3.5, with performance decaying from 86.86 \% to 75 \%.


\subsection{Negation Rejection Evaluation}

\begin{figure}
    \centering
    \includegraphics[width=0.99\linewidth]{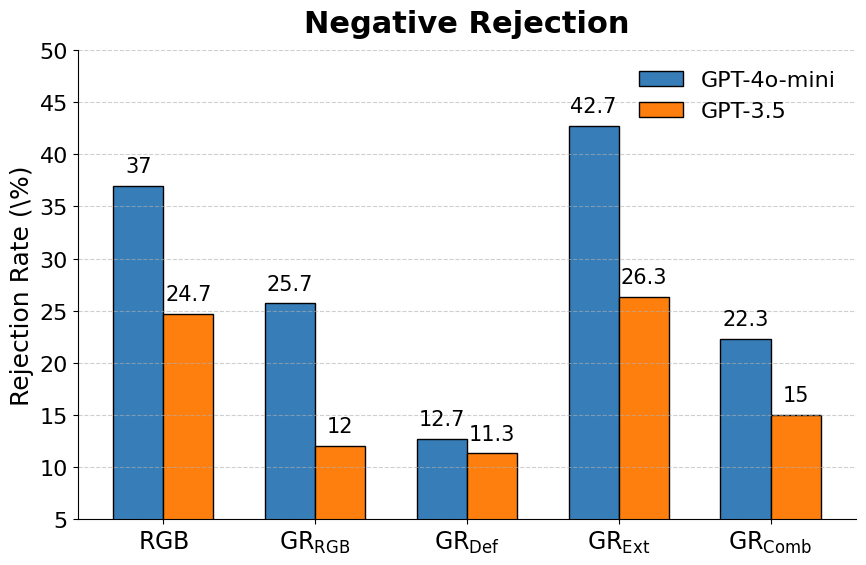}
    \caption{Comparative analysis for the negative rejection task measured by rejection rate (\%) for GPT-4o-mini and GPT-3.5.}
    \label{fig:negation}
\end{figure}

In this experiment, we assess the models' ability to recognize when the retrieved documents contain irrelevant information for answering the user's question. Thus, when both external sources and the LLM's encoded knowledge are insufficient, the model is expected to refrain from providing an answer. Figure \ref{fig:negation} summarizes the rejection rate attained by the five configurations considered in this study. Similar to the previous sections, the encoded knowledge within GPT-4o-mini plays a significant role in this experiments, confirming that the more complex model leads to better performance. The RGB baseline, which does not rely on structured knowledge, avoids answering questions when the information available is insufficient. All solutions based on knowledge graph underperformed the baseline, except for the GR$_{ext}$, which relies solely on the external knowledge. It produces the highest rejection rates for both models, i.e., 42.66\% for GPT-4o-mini and 33\% for GPT-3.5. This demonstrates that prompts that explicitly target document credibility and information adequacy can improve a models's performance in rejecting unanswerable questions. Moreover, it also shows that LLMs can be overconfident in their encoded knowledge, assuming that it can provide reliable responses in most scenarios. This is most evident for the configuration based on the GraphRAG default prompt, i.e., GR$_{def}$, which has the lowest rejection rate for GPT-4o-mini, 12.66\%, which demonstrates the model's propensity to be overconfident. Despite some variation across configurations, rejection rates are quite low (all below 50\%), indicating the overall challenge faced by current RAG systems in effectively rejecting unanswerable questions. These findings indicate the need for additional attention to the development of negative rejection approaches. Although external knowledge prompting (as in GR$_{ext}$) is promising, further refinement is necessary to improve the system's capacity to recognize when the retrieved information is not adequate to answer a particular question.

\section{Limitation and Future Work}
\label{sec:limitation}
While our approach demonstrates promising results, several directions remain to be explored. For instance, future research must include solutions based on KG as baselines. Additionally, we aim to investigate an unified solution for all four tasks addressed in the RGB benchmark. This would remove the need to customize the GraphRAG framework for each task studied in this work. For that, we will investigate hybrid approaches where structured knowledge will be combined with advanced reasoning mechanisms, such as chain-of-thoughts. In this context, the integration of confidence scores for retrieved facts can further refine the model capabilities, especially for the  negative rejection task. Finally, multimodal RAG is another direction to be explored to improve LLMs' reliability. These enhancements can solidify RAG systems as robust solutions for real-world deployments where information consistency and trustworthiness are critical.

\section{Conclusion}
\label{sec:conclusion}
In this study, we explore the combination of knowledge graph-based methods with Retrieval-Augmented Generation models. We show that by integrating these two approaches it is possible to enhance LLM robustness towards noisy and inconsistent retrieved information. We evaluate this approach in four tasks: noise robustness, information integration, negative rejection, and counterfactual robustness. The proposed customizations on the GraphRAG framework outperformed the vanilla RGB baseline, particularly for lower-complexity models like GPT-3.5. The experiments show that the combination of encoded and external knowledge, GR$_{comb}$, and the exclusive use of external knowledge, GR$_{ext}$, provide the greatest gains, with GR$_{ext}$ achieving the maximum rejection scores for unanswerable questions and GR$_{comb}$ achieving maximum error detection and correction performance. Interestingly, the increases in performance were greater in GPT-3.5 than in GPT-4o-mini, suggesting that knowledge graph augmentation is especially useful for models with limited encoded knowledge. While there are setbacks in adverse rejection scenarios where overall performance percentages remained below 50\%, our findings provide valuable insights into the construction of robust RAG systems that can handle inconsistencies in the retrieved information. 

\bibliographystyle{unsrt}
\bibliography{biblio.bib}

\newpage

\end{document}